\documentclass{INTERSPEECH2023}
 \interspeechcameraready

\title{Performance of data-driven inner speech decoding with same-task EEG-fMRI data fusion and bimodal models}
\name{Holly Wilson$^1$, Scott Wellington$^1$, Foteini Simistira Liwicki$^2$, Vibha Gupta$^2$, Rajkumar Saini$^2$, Kanjar De$^2$, Nosheen Abid$^2$, Sumit Rakesh$^2$, Johan Eriksson$^3$, Oliver Watts$^4$, Xi Chen$^1$, Mohammad Golbabaee$^1$, Michael J. Proulx$^1$, Marcus Liwicki$^2$, Eamonn O'Neill$^1$, Benjamin Metcalfe$^1$}
\address{
  $^1$University of Bath, UK \hspace{1cm} $^2$Luleå University of Technology, Sweden\\
  $^3$Umeå University, Sweden \hspace{3.54cm} $^4$SpeakUnique Ltd., UK
  }
\email{hlw69@bath.ac.uk, sdlw20@bath.ac.uk}

\begin{document}

\maketitle

\begin{abstract}
Decoding inner speech from the brain signal via hybridisation of fMRI and EEG data is explored to investigate the performance benefits over unimodal models. Two different bimodal fusion approaches are examined: concatenation of probability vectors output from unimodal fMRI and EEG machine learning models, and data fusion with feature engineering. Same-task inner speech data are recorded from four participants, and different processing strategies are compared and contrasted to previously-employed hybridisation methods. Data across participants are discovered to encode different underlying structures, which results in varying decoding performances between subject-dependent fusion models. Decoding performance is demonstrated as improved when pursuing bimodal fMRI-EEG fusion strategies, if the data show underlying structure.
\end{abstract}

\section{Introduction}

Since the inception and uptake of functional magnetic resonance imaging (fMRI) in the 90s and early 00s, much neurolinguistic research has been undertaken to discover the neural correlates of speech production and perception \cite{dong2018neuroscience}. A concurrent improvement to existing electroencephalography (EEG) technology has improved access and lowered cost, and neurolinguistic research has gradually shifted in its favour. However, for decoding inner and imagined speech, many EEG studies note the processing advantages that are gained from the signals offered by fMRI data (e.g. \cite{sereshkeh2017eeg,janssen2020exploring,gillis2021neural}). As such, a nascent interest in combining the two modalities has emerged: for speech and language processing, bimodal fusion methods have more recently been applied to paradigms for auditory decoding from oddball paradigms \cite{correa2009fusion,akhonda2018consecutive}, speech-gesture integration \cite{he2015eeg,he2018spatial}, and speech and language mapping \cite{ morillon2010neurophysiological,puschmann2017right}.

Both recording modalities have their own benefits: EEG has fine-grained millisecond temporal accuracy of up to 16,000 Hz, but limited spatial resolution. In contrast, fMRI has high  ~1mm spatial precision, but weak temporal resolution. Combining these two modalities can compensate for their individual limitations, and has been shown to improve decoding performance over unimodal decoding \cite{perronnet2017unimodal}.

Whilst fMRI has been used extensively to investigate the neural correlates of speech (e.g. \cite{belin2004thinking,fu2006fmri,christoffels2007neural}), it is only recently gaining traction for brain-computer interface (BCI)-motivated inner speech decoding. Using three subjects, Tang \textit{et al.} \cite{tang2022semantic} demonstrate how fMRI may successfully be used to reconstruct perceived continuous natural language based on its semantic representation; the decoder was used to subsequently identify, via pair-wise classification, which of five story transcripts the participant was imagining. Other support for fMRI-based language decoding comes from Correia \textit{et al.}'s study in which spoken words were shown to be decodable across languages based on semantic representations \cite{correia2014brain}.

EEG is also an increasingly-popular BCI tool for inner speech decoding; recently, van den Berg \textit{et al.} \cite{van2021inner} report a 35\% model accuracy with a 4-class inner speech decoding paradigm, while Kiroy \textit{et al.} \cite{kiroy2022spoken} report model accuracies from 33\% to 40\% with a 7-class inner speech task---notably both of these studies employed paradigms restricted to `spatial direction’ words. Decoding performances have also been shown to remain stable with commercial-grade dry electrode EEG devices which record the brain signal with lower fidelity and signal-to-noise ratio (SNR): Clayton \textit{et al.} \cite{clayton2020decoding} report a 69\% accuracy for a binary consonant-vowel classification task for an imagined speech paradigm, though model performances remained statistically significant for select participants who reported performing inner speech \textit{post hoc}.

Fusion can be simultaneous (i.e. EEG and fMRI record the brain signal concurrently), or non-simultaneous, in which same-task data is gathered independently for the two modalities at different times. Simultaneous fMRI-EEG recording provides different views of the same data, enabling temporal alignment of the two datasets. However, it also introduces artifacts, such as magnetic field distortions from the EEG electrodes \cite{uludaug2014general}, meaning data quality can be reduced and extra denoising required. In fMRI-EEG fusion, increasing the SNR is crucial for exploring optimal data representation and fusion strategies. In this context, non-simultaneous recording is a valuable alternative: not only is the ideal BCI robust across subjects, but also across the same task, recorded at different time-points. 

Fusion of non-simultaneous EEG and fMRI has been shown to boost classification performances over unimodal classification between 8 inner speech words \cite{liwicki2022bimodaldecode}. Results were significantly improved, with an average of 30.29\% ±2.71\% accuracy for 8-word classification, compared to ~18\% for fMRI, and ~22\% for EEG. Data can be fused at different stages: early, intermediate or late. In Simistira Liwicki \textit{et al.} \cite{liwicki2022bimodaldecode}, late fusion is evoked via an ensemble classification paradigm: probability vector predictions from EEG and fMRI subj-models are concatenated, and then input to a random forest (RF) classifier.

Whilst late fusion is promising for hybrid fMRI-EEG systems for inner speech, there are likely interdependencies between the two data types that can be exploited for improved decoding \cite{stahlschmidt2022multimodal}. Early fusion and intermediate fusion enable such cross-modal relationships to be found. Further, finding the optimal data representation of each view is another criterion for building successful hybrid systems. High dimensionality and low SNR can limit a classification model from discriminating between classes. In the research presented here, we explore and compare model performances between early and late fusion. With limited training data per class, we also use data augmentation to explore the data-driven performance of our models. 

\begin{figure}[t!]
  \centering
  \includegraphics[width=8cm]{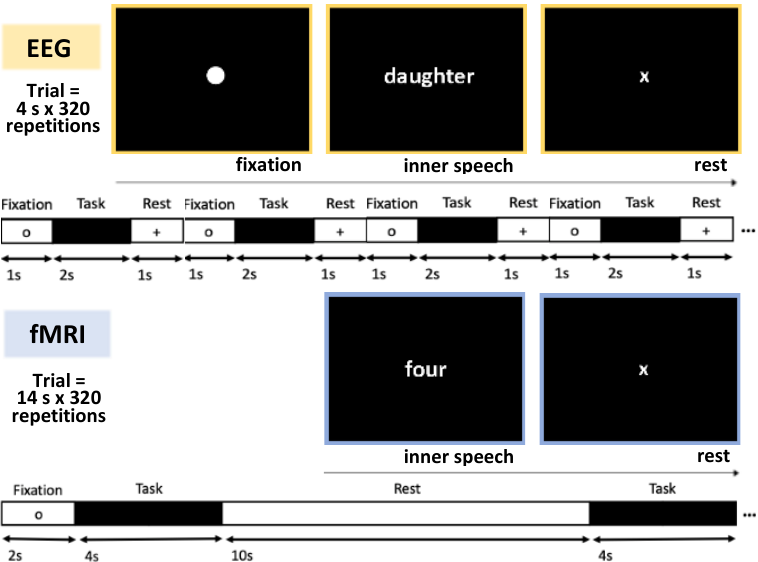}
  \caption{An illustration of the iSpeech protocol: in our same-task protocol, participants performed the same inner speech paradigm within independent EEG and fMRI recordings.}
  \label{fig:protocol}
\end{figure}

\section{iSpeech protocol}

The dataset employed by this study (iSpeech), is a publicly-available dataset shared by Simistira Liwicki \textit{et al.} \cite{liwicki2022bimodal-nature}. The words selected included four words each from two categories, numbers: \textit{four}, \textit{three}, \textit{ten} \textit{six}; and social: \textit{daughter}, \textit{father}, \textit{wife}, \textit{child} based on the research by Huth \textit{et al.} \cite{huth2016natural}.

The EEG experimental paradigm (see Figure \ref{fig:protocol}) involved one session with 40 trials for each of the eight words. A trial consisted of a 1-second fixation point, followed by 2 seconds of inner speech cued by the word stimuli presented on a screen, and a rest of 1 second. fMRI data were gathered in two sessions, with 40 trials for each word across the two sessions. In a single trial, the word was again cued on the screen, but 4 seconds for the inner speech task, followed by a 10-second rest. This longer rest time is important for the lag in the fMRI blood-oxygen-level-dependent (BOLD) signal to return to baseline. 

A BioSemi Active2 system with 64 electrodes was used to gather EEG data, with a sampling rate of 512 Hz, and 16-bit resolution. Impedances were kept between -20 k$\Omega$ and 20 k$\Omega$. For fMRI, a Siemens Magnetom Prisma MRI system was used. For functional images, TR = 2.16 s, voxel size = 2 $\times$ 2 $\times$ 2 mm, and for anatomical, a sagittal T1-weighted, TR = 662.0 and voxel size 3 $\times$ 3 $\times$ 2 mm. Technical validation of this data, alongside further acquisition details can be found in the data descriptor paper \cite{liwicki2022bimodal-nature}.

\section{Methodology}

\begin{figure}[b!]
  \hspace{-.5em}
  \includegraphics[width=8.2cm]{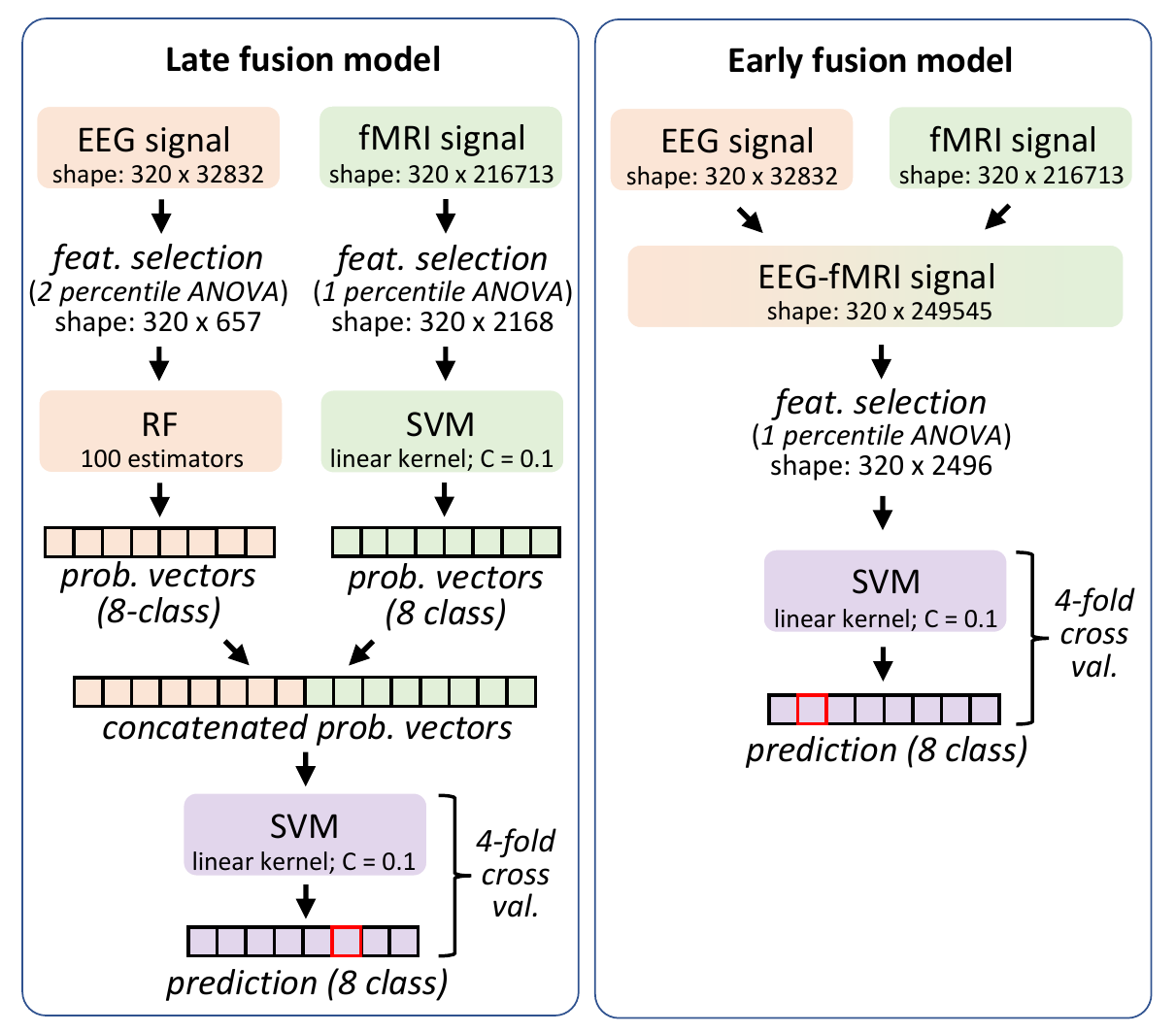}
  \caption{Illustration of the processing pipelines for 'late fusion' and 'early fusion' bimodal models. In the late fusion model, probability vectors from each submodel are concatenated and fed into an SVM classifier. In the early fusion model, data features rather than model outputs are concatenated. }
  \label{fig:methods}
\end{figure}

\begin{figure}[h]
  \hspace{-1.5em}
  \includegraphics[width=8.5cm]{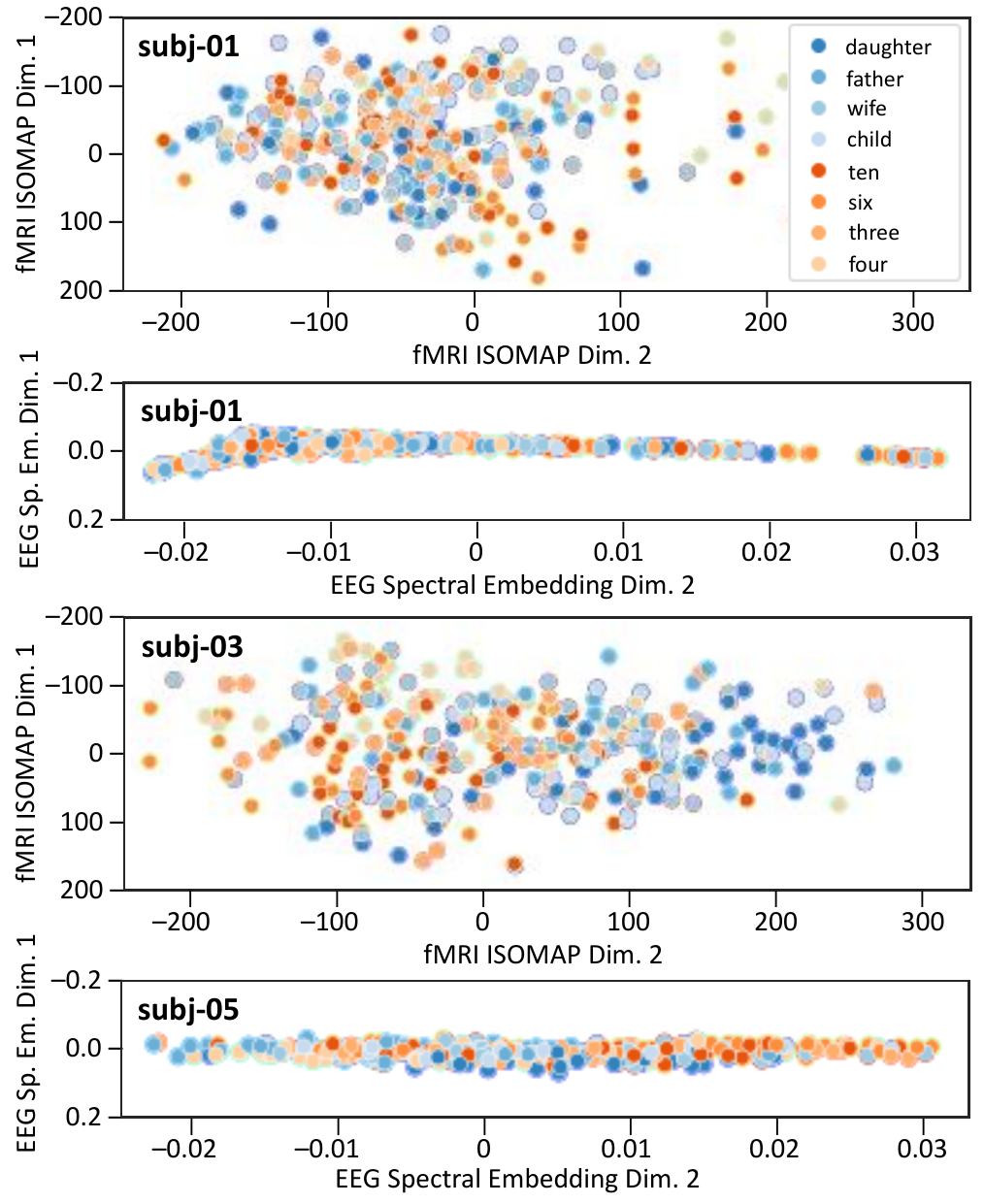}
  \caption{EEG and fMRI data from \textit{subj-03}, \textit{subj-01} and \textit{subj-05} projected onto a two-dimensional manifold space using isometric mapping and spectral embedding, to visualise the underlying structure in relation to our eight classes.}
  \label{fig:manifold}
\end{figure}

\subsection{\normalfont{\textit{EEG and fMRI preprocessing}}}

EEG data were referenced to two external electrodes placed on the mastoid processes, before applying a 1 Hz high-pass filter to remove DC drift. Line noise was removed with notch filters at 50 Hz and all harmonics up to the Nyquist frequency. ICA was applied to the data, and the highest-ranked component best describing electro-oculogram (EOG) artifacts was removed from all data. To prevent double-filtering the data with subsequent bandpass operations, the ICA weights calculated from this sanitised data are used to decompose a copy of the EEG data at the point of re-referencing, followed by the remaining cleaning operations.

fMRI data were motion-corrected after calculating spatial displacement maps, slice-time corrected and then co-registered to the T1-weighted structural scan. The data were then normalised to a canonical template, specifically Montreal Neurological Institute (MNI) space. To acquire beta images for classification, we estimated a single-trial based general linear model (GLM), convolved with a canonical haemodynamic response function. Movement parameters were used as additional regressors. Smoothing using a 8mm Gaussian kernel was applied, and background information removed using masking. 

\subsection{\normalfont{\textit{Classification Approaches}}}
Classification is performed for each subject separately. All cross validation is 4-fold resulting in 480 train exemplars and 160 test for each fold, with the exception of the augmented version, with 8720 training and 2906 test respectively. Augmentation involved shuffling the EEG and fMRI data within the training set, to make new pairings between the EEG and fMRI labels. See Figure \ref{fig:methods} for a visual depiction of the methodologies. Taking fMRI beta images flattens the time dimension, resulting in a single volumetric image per exemplar, which is flattened using masking into 320 $\times$ 216713 matrices. All code---including the fixed seeds with which experiments were run---will be made available on the lead author's GitHub repository for replicability.
\subsubsection{Unimodal Classification}
For \textit{fMRI unimoda}l classification, the top 1 percentile of discriminative features were selected using ANOVA. The data was standardised and then input to the  Support Vector Machine (SVM) model \cite{cortes1995support}, for which a grid search strategy was used to select the hyperparameters of C = 0.1 (the penalty value which sets the model's tolerance for misclassification) using a linear kernel.  \textit{EEG unimodal} classification was equivalent except, with the top 2 percentile of features and no scaling applied as the classifier used was a random forest classifier \cite{ho1995random} with 100 estimators (where each estimator is a decision tree classifier fit to subj-samples of the data using bootstrapping).
Deep learning models were created for unimodal decoding. fMRI 3D-CNN was composed of five 3D convolutional layers, with a final fully connected layer. The convolutional layers use dropout (0.3) and maxpooling followed by ReLU activation. ADAM optimizer was selected with cross entropy loss function and a learning rate of 0.0005. EEGNet \cite{lawhern2018eegnet} was used for unimodal EEG.

\subsubsection{Bimodal Classification}
In the current study, we employed both an early and a late fusion approach to combine the two modalities. For \textit{late fusion}, we use the unimodal classification but probability vectors for each subj-model are outputted rather than prediction vectors. The probability vectors are concatenated and fed into a linear SVM.  In the \textit{early fusion} approach, the EEG and fMRI data features are concatenated, and subsequently one percentile of features are selected on the joint data representation. Grid search was used to establish the best-performing SVM model used a linear kernel, with parameter C = 0.1. In the early fusion augmentation version, 15 new EEG and fMRI pairs are created for each label instance independently for the train and test sets, prior to the concatenation step.

\subsection{\normalfont{\textit{Non-linear Dimensionality Reduction}}}

 Non-linear dimensionality reduction was applied to each subject and modality to investigate the underlying structure relating to the eight classes. Strategies to project the data into a two dimensional embedding space, included isometric mapping, spectral embedding and local linear embedding.

\section{Results}

\begin{figure}[b!]
  \hspace{-1.5em}
  \includegraphics[width=8.5cm]{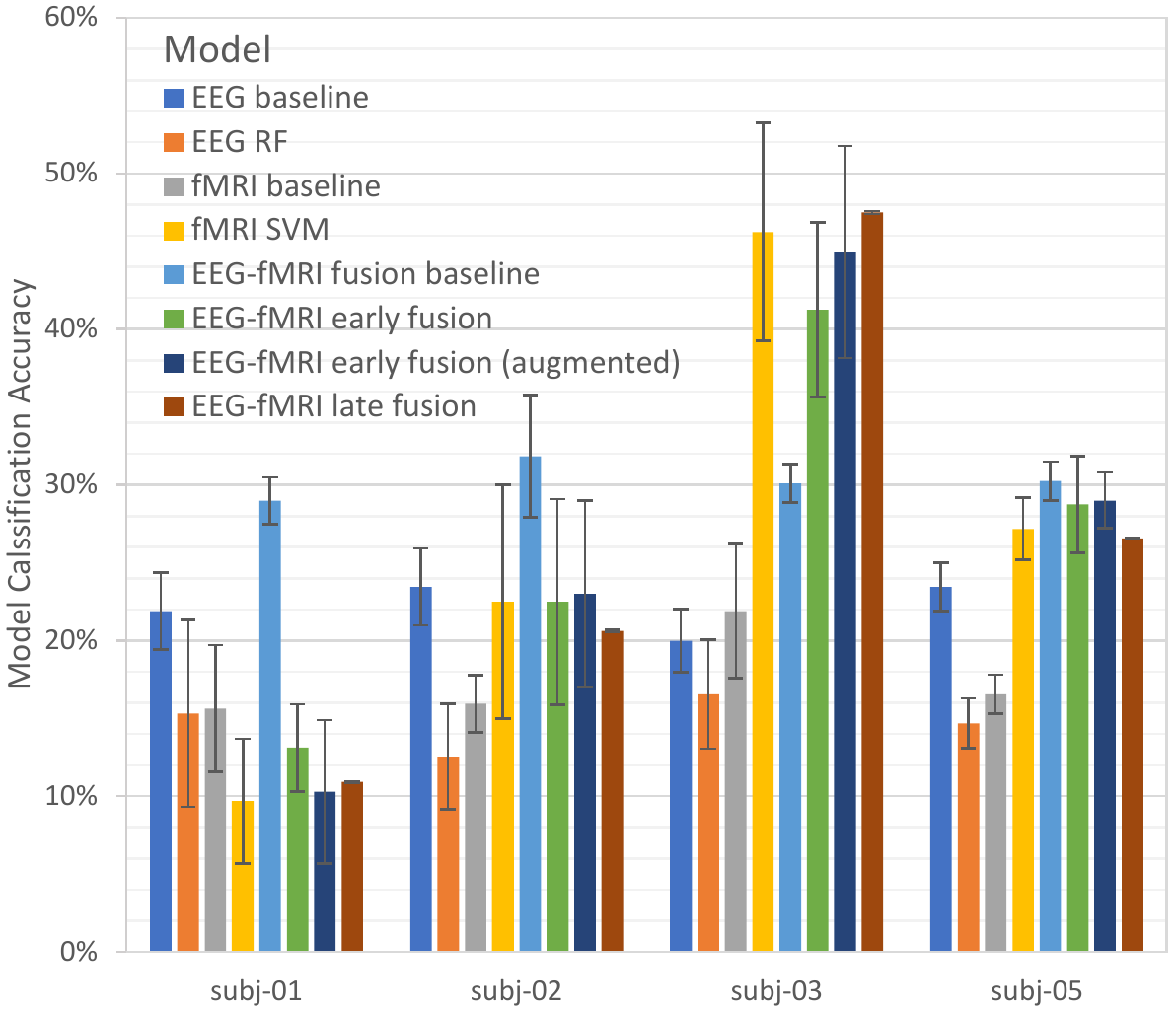}
  \caption{Classification accuracy for 8-class inner speech decoding using different data views and fusion approaches. Chance accuracy is 12.5\%. Baseline results refer to those obtained by Simistira Liwicki \textit{et al.} \cite{liwicki2022bimodaldecode}.}
  \label{fig:example2}
\end{figure}

\subsection{\normalfont{\textit{Manifold Structure}}}

Figure \ref{fig:manifold} visualises the underlying structure of the neural data after projection into a manifold space using isometric mapping and spectral embedding. For the fMRI data, the clearest structure is for \textit{subj-03} and \textit{subj-05}. Almost no structure is identifiable for \textit{subj-01} for either fMRI or EEG data. For EEG, \textit{subj-05} is the only subject to have visible structure.

\subsection{\normalfont{\textit{Unimodal Classification}}}

Our improved data representation of the fMRI data increased performance above the 21.88\% and 16.56\% classification baseline for \textit{subj-03} and \textit{subj-05} with decoding accuracies of 46.24\% and 27.18\% when using an SVM classifier with a linear kernel. Decoding on \textit{subj-01}'s fMRI data did not perform above chance level. For EEG, using an RF classifier, \textit{subj-01}, \textit{subj-03}, and \textit{subj-05} were decodable just above chance (16-17\%). See Table \ref{tab:classification}.
After 500 epochs of training and validation,  fMRI 3D-CNN achieved above chance on the test set, though worse than the SVM model, except with \textit{subj-01}. EEG-NET did not classify above baseline for any of the subjects.

\subsection{\normalfont{\textit{Bimodal Fusion}}}

For \textit{subj-03}, late fusion with improved data representation led to an improved classification accuracy on the fusion baseline, from 30.11\% to 44.96\% respectively. No performance enhancement was found for the remaining subjects (see Table \ref{tab:classification}). Adding augmentation to early fusion improved results marginally for \textit{subj-3}, but decreased results for \textit{subj-1}. We suggest that the low data quality (see Figure \ref{fig:manifold}) of \textit{subj-01} resulted in augmentation efforts lowering the SNR by increasing the presence of noise.

\begin{table}[th]
\caption{Unimodal and bimodal fusion model accuracies (\%).}
\label{tab:classification}
\centering
\begin{tabular}{@{}lrrrrr@{}}
\toprule
\textbf{Subject} & \textbf{sub-01} & \textbf{sub-02} & \textbf{sub-03} & \textbf{sub-05} & \textbf{Avg.}  \\ 
\midrule

EEG b.line &21.88  & 23.44 &20.00  & 23.44&22.19  \\ 
\textit{ std.} & 2.47 & 2.47 & 2.04 & 1.56 & 2.14\\

EEG RF & 15.31  & 12.55  &  16.56 & 14.68 & 14.78 \\ 
\textit{ std.} & 6.00 & 3.40 &  3.50 & 1.6& 3.63 \\ 

fMRI b.line & 15.63 &15.94  &21.88  &16.56 & 17.50 \\ 
\textit{ std.} & 4.07 & 1.82& 4.31 & 1.25& 2.86\\

fMRI SVM & 9.69 & 22.50 & 46.24 & 27.18 & 26.40\\
\textit{ std.} & 4.00 & 7.50 & 7.00 & 2.00 & 5.13 \\

\midrule
\midrule

Fusion b.line & 28.98 &31.84  &30.11   & 30.25&30.29 \\ 
\textit{ std.} & 1.49 &3.94& 1.24 & 1.24& 2.71\\

Early fusion & 13.12 & 22.49 &  41.25 & 28.75&26.40 \\ 
\textit{ std.} & 2.80 &6.60& 5.60 & 3.10&4.53  \\

Early f. (aug) &10.29  & 22.99 &44.96   & 29.00& 26.81\\ 
\textit{ std.} & 4.60 &6.00& 6.80 &1.80& 4.8\\

Late fusion & 10.93 & 20.63 & 47.50  & 26.56& 26.41\\ 
\textit{ std.} & 4.00 &5.70&9.10  & 1.00& 4.95 \\

%\midrule
%\midrule
\bottomrule
\end{tabular}
\end{table}

\section{Discussion}
Decoding inner speech from brain signals offers an exciting but challenging pursuit. fMRI-EEG fusion is proposed as a beneficial strategy for boosting decoding performance due to the distinct views that the modalities offer on inner speech processing.
 We built on our previous work fusing EEG and fMRI data for inner speech decoding \cite{liwicki2022bimodaldecode} to explore how choices made for enhanced data representation strategy, data augmentation and data fusion, impact classification success.
 
The improvements made to performance were inconsistent across subjects. Namely, \textit{subj-05} and \textit{subj-03} benefited from the new fMRI data representation and \textit{subj-05} from fusion strategies. Additionally, unimodal EEG decoding achieved barely above chance level. On a surface level these results are discouraging, and certainly from just the decoding performance, no robust recommendation can be made to use one fusion strategy over another or indeed to use fusion at all. To shed light on what might be driving the variance in improvement between subjects, we investigated the underlying data structure of EEG and fMRI for each subject. To do so, we applied non-linear dimensionality reduction via isometric mapping and spectral embedding, which enables class clusters and separation to be visualised. The findings from each subject's visualisation align with, and help explain,  their respective decoding results.  Structure is visible for all subjects' fMRI data, except \textit{subj-01}. This is consistent with unimodal fMRI decoding performance: \textit{subj-01} scored below chance, whilst \textit{subj-02}, \textit{03}, and \textit{05} ranged from 23--46\% accuracy. Conversely, coherent visible structure for EEG data was present in just \textit{subj-01} and \textit{05}. Remarkably, \textit{subj-05} was the only subject for which structure is visible in both the EEG and fMRI data, and for which fusion notably boosted performance beyond unimodal fMRI. In comparison, \textit{subj-01} shows no meaningful structure for the fMRI data and very little for the EEG data.

Using improved processing techniques, the present study achieved an average of 26\% decoding accuracy on fMRI data, surpassing the 18\% baseline result in Liwicki \textit{et al.} \cite{liwicki2022bimodaldecode}. This indicates the promise of using feature selection and preprocessing techniques that reduce the data dimensionality, such as applying a mask to remove background information surrounding the brain and taking only a top percentile of discrimiative features. Lower dimensionality enables complex decoding challenges to be solved with simpler models, as demonstrated with the SVM model used in this work. Further, smoothing the fMRI data can reduce uncorrelated random noise in voxels and increase SNR \cite{lindquist2008statistical}. fMRI exhibits potential in hybrid BCIs, particularly in capturing semantic-based representations, as evidenced by the manifold visualisations. Though likely, there is further information such as phonological representation that can be extracted and exploited in fMRI.

Using deep learning to decode inner and imagined speech has revealed mixed results \cite{van2021inner,cooney2018mel}. In this preliminary study, both the number of participants and trials for each word were insufficient for more powerful methods such as deep learning and intermediate fusion approaches to have any benefit, as demonstrated by the poor decoding performance of both our EEG and fMRI deep learning models. Interestingly, van den Berg \textit{et al.} \cite{van2021inner} report low deep learning decoding performances for all but one subject, for which they obtained 34.5\% accuracy on an 8-class problem. This highlights the variance between subjects' decodability. We recommend assessing the underlying structure of each subject before classification. Gathering large datasets for such tasks is a challenge, but will be paramount for conclusive evaluations of complementary data in hybrid systems.

In this study we fused intra-subject data. The idea of combining different but related views of data can be extended. Fusing same-task inter-subject data boosts the dataset size and facilitates generalisable BCIs. Anatomical and functional alignment are well-demonstrated approaches to transform inter-subject fMRI data into a common space, whilst maintaining the underlying data structure \cite{haxby2011common}. For EEG, inter-subject fusion is less explored (though see \cite{arevalillo2019combining} for inter-subject non-linear normalisation). However, if a common underlying structure can be discovered between fMRI and EEG that is consistent across subjects, then a hybrid bimodal and inter-subject fusion could potentially be applied. Again, this necessitates the creation of large datasets.

\section{Conclusion}
For an inner speech decoding task, we assessed the performance benefits of hybrid fMRI-EEG over unimodal models. Subject-specific decoding performance aligned with visualisations of the underlying data structure. This indicates that data with a more apparent structure may benefit from fusion techniques. Interestingly, fMRI showed higher unimodal decoding performance than EEG for two subjects, supporting fMRI's efficacy for inner speech decoding. Deep learning models performed poorly on this dataset, highlighting a need for larger datasets at both a subject and exemplar level. Such datasets will present exciting opportunities to exploit inter-subject and multimodality fusion for inner speech decoding.

\section{Acknowledgements}

This research was funded by: the Grants for Excellent Research Projects of SRT.ai 2022; the United Kingdom Research Institute (UKRI; grant EP/S023437/1); and the Engineering and Physical Sciences Research Council (EPSRC; grant EP/S515279/1).

\bibliographystyle{IEEEtran}
\bibliography{mybib}

\begin{thebibliography}{10}
\providecommand{\url}[1]{#1}
\csname url@samestyle\endcsname
\providecommand{\newblock}{\relax}
\providecommand{\bibinfo}[2]{#2}
\providecommand{\BIBentrySTDinterwordspacing}{\spaceskip=0pt\relax}
\providecommand{\BIBentryALTinterwordstretchfactor}{4}
\providecommand{\BIBentryALTinterwordspacing}{\spaceskip=\fontdimen2\font plus
\BIBentryALTinterwordstretchfactor\fontdimen3\font minus
  \fontdimen4\font\relax}
\providecommand{\BIBforeignlanguage}[2]{{%
\expandafter\ifx\csname l@#1\endcsname\relax
\typeout{** WARNING: IEEEtran.bst: No hyphenation pattern has been}%
\typeout{** loaded for the language `#1'. Using the pattern for}%
\typeout{** the default language instead.}%
\else
\language=\csname l@#1\endcsname
\fi
#2}}
\providecommand{\BIBdecl}{\relax}
\BIBdecl

\bibitem{dong2018neuroscience}
L.~Dong, C.~Luo, X.~Liu, S.~Jiang, F.~Li, H.~Feng, J.~Li, D.~Gong, and D.~Yao,
  ``Neuroscience information toolbox: An open source toolbox for eeg--fmri
  multimodal fusion analysis,'' \emph{Frontiers in Neuroinformatics}, vol.~12,
  p.~56, 2018.

\bibitem{sereshkeh2017eeg}
A.~R. Sereshkeh, R.~Trott, A.~Bricout, and T.~Chau, ``Eeg classification of
  covert speech using regularized neural networks,'' \emph{IEEE/ACM
  Transactions on Audio, Speech, and Language Processing}, vol.~25, no.~12, pp.
  2292--2300, 2017.

\bibitem{janssen2020exploring}
N.~Janssen, M.~v.~d. Meij, P.~J. L{\'o}pez-P{\'e}rez, and H.~A. Barber,
  ``Exploring the temporal dynamics of speech production with eeg and group
  ica,'' \emph{Scientific reports}, vol.~10, no.~1, p. 3667, 2020.

\bibitem{gillis2021neural}
M.~Gillis, J.~Vanthornhout, J.~Z. Simon, T.~Francart, and C.~Brodbeck, ``Neural
  markers of speech comprehension: measuring eeg tracking of linguistic speech
  representations, controlling the speech acoustics,'' \emph{Journal of
  Neuroscience}, vol.~41, no.~50, pp. 10\,316--10\,329, 2021.

\bibitem{correa2009fusion}
N.~M. Correa, Y.-O. Li, T.~Adali, and V.~D. Calhoun, ``Fusion of fmri, smri,
  and eeg data using canonical correlation analysis,'' in \emph{2009 IEEE
  International Conference on Acoustics, Speech and Signal Processing}.\hskip
  1em plus 0.5em minus 0.4em\relax IEEE, 2009, pp. 385--388.

\bibitem{akhonda2018consecutive}
M.~A. Akhonda, Y.~Levin-Schwartz, S.~Bhinge, V.~D. Calhoun, and T.~Adali,
  ``Consecutive independence and correlation transform for multimodal fusion:
  Application to eeg and fmri data,'' in \emph{2018 IEEE International
  Conference on Acoustics, Speech and Signal Processing (ICASSP)}.\hskip 1em
  plus 0.5em minus 0.4em\relax IEEE, 2018, pp. 2311--2315.

\bibitem{he2015eeg}
Y.~He, H.~Gebhardt, M.~Steines, G.~Sammer, T.~Kircher, A.~Nagels, and
  B.~Straube, ``The eeg and fmri signatures of neural integration: An
  investigation of meaningful gestures and corresponding speech,''
  \emph{Neuropsychologia}, vol.~72, pp. 27--42, 2015.

\bibitem{he2018spatial}
Y.~He, M.~Steines, J.~Sommer, H.~Gebhardt, A.~Nagels, G.~Sammer, T.~Kircher,
  and B.~Straube, ``Spatial--temporal dynamics of gesture--speech integration:
  a simultaneous eeg-fmri study,'' \emph{Brain Structure and Function}, vol.
  223, pp. 3073--3089, 2018.

\bibitem{morillon2010neurophysiological}
B.~Morillon, K.~Lehongre, R.~S. Frackowiak, A.~Ducorps, A.~Kleinschmidt,
  D.~Poeppel, and A.-L. Giraud, ``Neurophysiological origin of human brain
  asymmetry for speech and language,'' \emph{Proceedings of the National
  Academy of Sciences}, vol. 107, no.~43, pp. 18\,688--18\,693, 2010.

\bibitem{puschmann2017right}
S.~Puschmann, S.~Steinkamp, I.~Gillich, B.~Mirkovic, S.~Debener, and C.~M.
  Thiel, ``The right temporoparietal junction supports speech tracking during
  selective listening: Evidence from concurrent eeg-fmri,'' \emph{Journal of
  Neuroscience}, vol.~37, no.~47, pp. 11\,505--11\,516, 2017.

\bibitem{perronnet2017unimodal}
L.~Perronnet, A.~L{\'e}cuyer, M.~Mano, E.~Bannier, F.~Lotte, M.~Clerc, and
  C.~Barillot, ``Unimodal versus bimodal eeg-fmri neurofeedback of a motor
  imagery task,'' \emph{Frontiers in Human Neuroscience}, vol.~11, p. 193,
  2017.

\bibitem{belin2004thinking}
P.~Belin, S.~Fecteau, and C.~Bedard, ``Thinking the voice: neural correlates of
  voice perception,'' \emph{Trends in cognitive sciences}, vol.~8, no.~3, pp.
  129--135, 2004.

\bibitem{fu2006fmri}
C.~H. Fu, G.~N. Vythelingum, M.~J. Brammer, S.~C. Williams, E.~Amaro~Jr, C.~M.
  Andrew, L.~Y{\'a}g{\"u}ez, N.~E. Van~Haren, K.~Matsumoto, and P.~K. McGuire,
  ``An fmri study of verbal self-monitoring: neural correlates of auditory
  verbal feedback,'' \emph{Cerebral Cortex}, vol.~16, no.~7, pp. 969--977,
  2006.

\bibitem{christoffels2007neural}
I.~K. Christoffels, E.~Formisano, and N.~O. Schiller, ``Neural correlates of
  verbal feedback processing: an fmri study employing overt speech,''
  \emph{Human brain mapping}, vol.~28, no.~9, pp. 868--879, 2007.

\bibitem{tang2022semantic}
J.~Tang, A.~LeBel, S.~Jain, and A.~G. Huth, ``Semantic reconstruction of
  continuous language from non-invasive brain recordings,'' \emph{bioRxiv}, pp.
  2022--09, 2022.

\bibitem{correia2014brain}
J.~Correia, E.~Formisano, G.~Valente, L.~Hausfeld, B.~Jansma, and M.~Bonte,
  ``Brain-based translation: fmri decoding of spoken words in bilinguals
  reveals language-independent semantic representations in anterior temporal
  lobe,'' \emph{Journal of Neuroscience}, vol.~34, no.~1, pp. 332--338, 2014.

\bibitem{van2021inner}
B.~van~den Berg, S.~van Donkelaar, and M.~Alimardani, ``Inner speech
  classification using eeg signals: A deep learning approach,'' in \emph{2021
  IEEE 2nd International Conference on Human-Machine Systems (ICHMS)}.\hskip
  1em plus 0.5em minus 0.4em\relax IEEE, 2021, pp. 1--4.

\bibitem{kiroy2022spoken}
V.~N. Kiroy, O.~Bakhtin, E.~Krivko, D.~M. Lazurenko, E.~Aslanyan,
  D.~Shaposhnikov, and I.~V. Shcherban, ``Spoken and inner speech-related eeg
  connectivity in different spatial direction,'' \emph{Biomedical Signal
  Processing and Control}, vol.~71, p. 103224, 2022.

\bibitem{clayton2020decoding}
J.~Clayton, S.~Wellington, C.~Valentini-Botinhao, and O.~Watts, ``Decoding
  imagined, heard, and spoken speech: Classification and regression of eeg
  using a 14-channel dry-contact mobile headset.'' in \emph{INTERSPEECH}, 2020,
  pp. 4886--4890.

\bibitem{uludaug2014general}
K.~Uluda{\u{g}} and A.~Roebroeck, ``General overview on the merits of
  multimodal neuroimaging data fusion,'' \emph{Neuroimage}, vol. 102, pp.
  3--10, 2014.

\bibitem{liwicki2022bimodaldecode}
F.~S. Liwicki, V.~Gupta, R.~Saini, K.~De, N.~Abid, S.~Rakesh, S.~Wellington,
  H.~Wilson, M.~Liwicki, and J.~Eriksson, ``{Bimodal pilot study on inner
  speech decoding reveals the potential of combining EEG and fMRI},'' 2022.

\bibitem{stahlschmidt2022multimodal}
S.~R. Stahlschmidt, B.~Ulfenborg, and J.~Synnergren, ``Multimodal deep learning
  for biomedical data fusion: a review,'' \emph{Briefings in Bioinformatics},
  vol.~23, no.~2, p. bbab569, 2022.

\bibitem{liwicki2022bimodal-nature}
F.~S. Liwicki, V.~Gupta, R.~Saini, K.~De, N.~Abid, S.~Rakesh, S.~Wellington,
  H.~Wilson, M.~Liwicki, and J.~Eriksson, ``Bimodal
  electroencephalography-functional magnetic resonance imaging dataset for
  inner-speech recognition,'' \emph{bioRxiv}, pp. 2022--05, 2022.

\bibitem{huth2016natural}
A.~G. Huth \emph{et~al.}, ``Natural speech reveals the semantic maps that tile
  human cerebral cortex,'' \emph{Nature}, vol. 532, no. 7600, pp. 453--458,
  2016.

\bibitem{cortes1995support}
C.~Cortes and V.~Vapnik, ``Support-vector networks,'' \emph{Machine learning},
  vol.~20, pp. 273--297, 1995.

\bibitem{ho1995random}
T.~K. Ho, ``Random decision forests,'' in \emph{Proceedings of 3rd
  international conference on document analysis and recognition}, vol.~1.\hskip
  1em plus 0.5em minus 0.4em\relax IEEE, 1995, pp. 278--282.

\bibitem{lawhern2018eegnet}
V.~J. Lawhern, A.~J. Solon, N.~R. Waytowich, S.~M. Gordon, C.~P. Hung, and
  B.~J. Lance, ``Eegnet: a compact convolutional neural network for eeg-based
  brain--computer interfaces,'' \emph{Journal of neural engineering}, vol.~15,
  no.~5, p. 056013, 2018.

\bibitem{lindquist2008statistical}
M.~A. Lindquist, ``The statistical analysis of fmri data,'' \emph{Statistical
  Science}, vol.~23, no.~4, pp. 439--464, 2008.

\bibitem{cooney2018mel}
C.~Cooney, R.~Folli, and D.~Coyle, ``Mel frequency cepstral coefficients
  enhance imagined speech decoding accuracy from eeg,'' in \emph{2018 29th
  Irish Signals and Systems Conference (ISSC)}.\hskip 1em plus 0.5em minus
  0.4em\relax IEEE, 2018, pp. 1--7.

\bibitem{haxby2011common}
J.~V. Haxby, J.~S. Guntupalli, A.~C. Connolly, Y.~O. Halchenko, B.~R. Conroy,
  M.~I. Gobbini, M.~Hanke, and P.~J. Ramadge, ``A common, high-dimensional
  model of the representational space in human ventral temporal cortex,''
  \emph{Neuron}, vol.~72, no.~2, pp. 404--416, 2011.

\bibitem{arevalillo2019combining}
M.~Arevalillo-Herr{\'a}ez, M.~Cobos, S.~Roger, and M.~Garc{\'\i}a-Pineda,
  ``Combining inter-subject modeling with a subject-based data transformation
  to improve affect recognition from eeg signals,'' \emph{Sensors}, vol.~19,
  no.~13, p. 2999, 2019.

\end{thebibliography}

\end{document}